\documentclass[10pt,journal,compsoc]{IEEEtran}

\usepackage{graphicx}
\usepackage[fleqn]{amsmath}
\usepackage{algorithm}
\usepackage[labelsep=space]{caption}
\usepackage{setspace}
\usepackage{algpseudocode}
\usepackage{amssymb}
\usepackage{bm}
\usepackage{color}
\usepackage{colortbl}
\usepackage{amsmath}
\usepackage{amssymb}
\usepackage{array}
\usepackage{booktabs}
\usepackage{hyperref}
\usepackage[tight,footnotesize]{subfigure}
\usepackage{enumerate}
\usepackage{mathtools,cuted}
\usepackage{stfloats}
\usepackage{adjustbox}
\usepackage{cite}
\usepackage{subfigure}
\usepackage{float}
\usepackage{multirow}
\usepackage{caption}
\usepackage{caption}
\usepackage{url}
\usepackage{leftidx}
\usepackage{setspace}
\ifCLASSINFOpdf
\else
\fi
\begin{document}
\title{NDPNet: A novel non-linear data projection network for few-shot fine-grained image classification}

\author{Weichuan Zhang$^{\dagger}$, \textsl{Member}, \textsl{IEEE}, Xuefang Liu$^{\dagger}$, Zhe Xue, Yongsheng Gao, \textsl{Senior Member}, \textsl{IEEE}, Changming Sun

%\author{Weichuan Zhang\textsuperscript{^{*}}, \textsl{Member}, \textsl{IEEE}, Jiapan Guo\textsuperscript{^{*}}, Yongsheng Gao, \textsl{Senior Member}, \textsl{IEEE}, Changming Sun

\IEEEcompsocitemizethanks{\IEEEcompsocthanksitem $^{\dagger}$ These authors contributed equally.\protect

%\IEEEcompsocthanksitem W. Zhang is with the Institute for Integrated and Intelligent Systems, Griffith University, QLD, Australia, and also with CSIRO Data61, PO Box 76, Epping, NSW 1710, Australia.\protect\\
%E-mail: zwc2003@163.com
%\IEEEcompsocthanksitem J. Guo is with the Department of Radiation Oncology, University Medical Center Groningen, University of Groningen, Hanzeplein 1, 9713GZ, Groningen, Netherlands.\protect\\
%E-mail: j.guo@rug.nl
%\IEEEcompsocthanksitem Y. Gao is with the Institute for Integrated and Intelligent Systems, Griffith University, QLD, Australia.\protect\\
%E-mail: yongsheng.gao@griffith.edu.au
%\IEEEcompsocthanksitem C. Sun is with CSIRO Data61, PO Box 76, Epping, NSW 1710, Australia.\protect\\
%E-mail: changming.sun@csiro.au

\IEEEcompsocthanksitem W. Zhang is with the Institute for Integrated and Intelligent Systems, Griffith University, QLD, Australia.\protect\\
E-mail: zwc2003@163.com
\IEEEcompsocthanksitem X. Liu is with Xidian University, Xi'an City, Shaanxi Province, China.\protect\\
E-mail: xfliu1@mail.xidian.edu.cn
\IEEEcompsocthanksitem Z. Xue is with Xidian University, Xi'an City, Shaanxi Province, China.\protect\\
E-mail: 1628882365@qq.com
\IEEEcompsocthanksitem Y. Gao is with the Institute for Integrated and Intelligent Systems, Griffith University, QLD, Australia.\protect\\
E-mail: yongsheng.gao@griffith.edu.au
\IEEEcompsocthanksitem C. Sun is with CSIRO Data61, PO Box 76, Epping, NSW 1710, Australia.\protect\\
E-mail: changming.sun@csiro.au

}
}

%\markboth{IEEE Transactions on Pattern Analysis and Machine Intelligence, April~2021}%
%{Shell \MakeLowercase{\textit{et al.}}: Bare Advanced Demo of IEEEtran.cls for IEEE Computer Society Journals}

\IEEEtitleabstractindextext{%
\begin{abstract}
Metric-based few-shot fine-grained image classification (FSFGIC) aims to learn a transferable feature embedding network by estimating the similarities between query images and support classes from very few examples. In this work, we propose, for the first time, to introduce the non-linear data projection concept into the design of FSFGIC architecture in order to address the limited sample problem in few-shot learning and at the same time to increase the discriminability of the model for fine-grained image classification. Specifically, we first design a feature re-abstraction embedding network that has the ability to not only obtain the required semantic features for effective metric learning but also re-enhance such features with finer details from input images. Then the descriptors of the query images and the support classes are projected into different non-linear spaces in our proposed similarity metric learning network to learn discriminative projection factors. This design can effectively operate in the challenging and restricted condition of a FSFGIC task for making the distance between the samples within the same class smaller and the distance between samples from different classes larger and for reducing the coupling relationship between samples from different categories. Furthermore, a novel similarity measure based on the proposed non-linear data project is presented for evaluating the relationships of feature information between a query image and a support set. It is worth to note that our proposed architecture can be easily embedded into any episodic training mechanisms for end-to-end training from scratch. Extensive experiments on FSFGIC tasks demonstrate the superiority of the proposed methods over the state-of-the-art benchmarks.
\end{abstract}

}

\maketitle

\IEEEdisplaynontitleabstractindextext

\IEEEpeerreviewmaketitle

\ifCLASSOPTIONcompsoc
\IEEEraisesectionheading{\section{Introduction}\label{sec:introduction}}
\else
\section{Introduction}
\label{sec:introduction}
\fi

\IEEEPARstart{H}{umans} are capable of learning stable feature representations with small training samples for dealing with image classification tasks~\cite{schmidt2009meaning}. Inspired by this ability of human, few-shot learning has attracted much attention in computer vision and pattern recognition which intends to rapidly learn a classifier with good generalization capacity for understanding new concepts from very limited numbers of labeled training examples.

Currently, the main problem of few-shot learning~\cite{li2019distribution,ijcai2020-100} is the very limited training examples of each class to effectively express a concept. In recent years, different few-shot learning methods~\cite{snell2017prototypical,wei2019piecewise,sung2018learning,li2020revisiting,zhu2020multi} have been presented to tackle fine-grained image classification tasks by learning transferable knowledge~\cite{snell2017prototypical} from an additional auxiliary dataset. The existing few-shot fine-grained image classification (FSFGIC) methods can be roughly classified into two main streams:~meta-learning based methods~\cite{wei2019piecewise,zhu2020multi} and metric-learning based methods~\cite{snell2017prototypical,li2019distribution,zhang2020deepemd}. Meta-learning based methods aim to learn meta knowledge with only a few training examples for each category. A set of functions are used for mapping labeled training image samples and test samples for visual classification. Metric-learning based methods intend to learn a group of functions for transforming test samples into the embedding space. And then the test samples will be classified into a class by a given similarity measure (e.g., nearest neighbor~\cite{snell2017prototypical} and cosine metric~\cite{li2020revisiting}).

In this work, our interest lies in addressing the metric-learning problem for FSFGIC tasks. One of the key issues of the metric-based FSFGIC methods is how to utilize a convolution neural network~\cite{vinyals2016matching, krizhevsky2017imagenet} and a similarity measure to learn a common feature representation for each category~\cite{snell2017prototypical}. Generally, mainstream metric-based FSFGIC methods follow the episodic training paradigm~\cite{vinyals2016matching}, and then image level feature representations~\cite{8658920} or a group of local feature descriptors~\cite{ijcai2020-100} from each input image are obtained for measuring the similarities~\cite{sung2018learning} between query images and support classes. Although the aforementioned methods~\cite{snell2017prototypical,sung2018learning,8658920,li2020revisiting,ijcai2020-100} have achieved a certain degree of success, given that the sample data in FSFGIC is extremely limited and the similarity between different categories in FSFGIC may be very high, we argue that projecting samples from query images and samples from support sets into a non-linear space with different non-linear projection factors and letting the designed network learn appropriate non-linear projection factors can make the distance between samples within a category smaller and the distance between samples from different categories larger and thus can effectively improve the classification performance in FSFGIC.

Currently, four convolutional blocks (i.e.,~\textit{Conv-64F}~\cite{vinyals2016matching}) are widely used in meta-learning based FSFGIC tasks~\cite{finn2017model, munkhdalai2018rapid, zhu2020multi} and metric-learning based FSFGIC tasks~\cite{li2020revisiting, huang2020low, ijcai2020-100}. Feature structure maps~\cite{finn2017model, munkhdalai2018rapid, zhu2020multi, li2020revisiting, huang2020low, ijcai2020-100} extracted from an input image with a single scale are used as descriptors for calculating the relationships between query images and support sets and performing FSFGIC. They follow the idea~\cite{girshick2015fast, ren2016faster} that descriptors extracted on a single scale image have a good trade-off between classification accuracy and speed. Our research indicates that they utilized the descriptors with strong semantic information for FSFGIC, but ignored the useful information of finer details in an image.

In this paper, we propose to formalize the FSFGIC as a feature distribution problem and design a novel non-linear data projection based metric learning network (NDPNet) for tackling the classification problems of FSFGIC. Firstly, we design a novel feature re-abstraction embedding network that has the ability to extract descriptors with detail re-enhanced semantic information from input images with a single scale. Meanwhile, a group of feature descriptors with detail re-enhanced semantic information are used to represent each test image. Secondly, a novel non-linear data projection strategy is designed to make the distance between the samples within the same class smaller and the distance between samples of different classes larger. Thirdly, a novel non-linear data projection based similarly measure is designed for obtaining the relationships between query samples after non-linear projection and support classes after non-linear projection. Finally, the Adam optimization method~\cite{kingma2014adam} with a cross-entropy loss is employed to train the whole network for learning the weights (including appropriate non-linear projection factors) and performing FSFGIC tasks.

The main contributions of our proposed method are as follows:

$\bullet$ A novel feature re-abstraction embedding network is designed which is capable of representing a sample by a set of feature representations with detail re-enhanced semantic information from an input image.

$\bullet$ A novel non-linear data projection strategy is designed for making the distance between the samples within the same class smaller and the distance between samples from different classes larger.

$\bullet$ A novel non-linear data projection based metric learning network is designed to learn appropriate non-linear projection factors for performing FSFGIC tasks.

Experiments on four fine-gained image classification benchmark datasets exhibit that our proposed method outperforms the state-of-the-art methods on various FSFGIC tasks.

\section{Related Works}
The existing FSFGIC methods can be roughly classified into two groups: meta-learning based FSFGIC methods and metric-learning based FSFGIC methods. We briefly review the two main streams in the FSFGIC literature as follows.

{\bf Meta-learning based FSFGIC methods.} The meta-learning based FSFGIC methods~\cite{thrun1998lifelong,thrun1998learning} aim to train a meta-learner which learns how to update the parameters of a given initial model with only a few training examples for each category. In order to more effectively distinguish some subtle and local differences between different fine-grained categories, recent meta-learning works have been presented for tackling FSFGIC tasks by considering how to more effectively capture subtle differences from limited training samples such as a piecewise mapping strategy~\cite{wei2019piecewise} and a multi-attention mechanism~\cite{zhu2020multi}.

Although the existing meta-learning based methods have achieved competitive results for FSFGIC, it is a difficult task to train a sophisticated memory addressing framework because of the temporally-linear hidden state dependencies~\cite{mishra2017simple}. In contrast, the proposed NDPNet architectures can be easily embedded into the episodic training mechanism for end-to-end training from scratch.

{\bf Metric-learning based FSFGIC methods.} The metric-learning based FSFGIC methods~\cite{vinyals2016matching,snell2017prototypical,li2019distribution,zhang2020deepemd,li2020revisiting,zhang2020deepemd,ijcai2020-100} aim to learn a transferable feature knowledge and obtain a distribution based on similarity metrics between different categories. Inspired by the Siamese neural network~\cite{koch2015siamese} and the episodic training mechanism~\cite{vinyals2016matching}, Prototypical-Net~\cite{snell2017prototypical} was proposed for few-shot learning tasks. In~\cite{snell2017prototypical}, a prototype was designed as the mean of embedded support examples for each class, and then Euclidean distance is used as the metric measure for FSFGIC.

\begin{figure*}[!htbp]
\setlength{\abovecaptionskip}{5pt}
\setlength{\belowcaptionskip}{5pt}
\centering
\includegraphics[width=2.1\columnwidth]{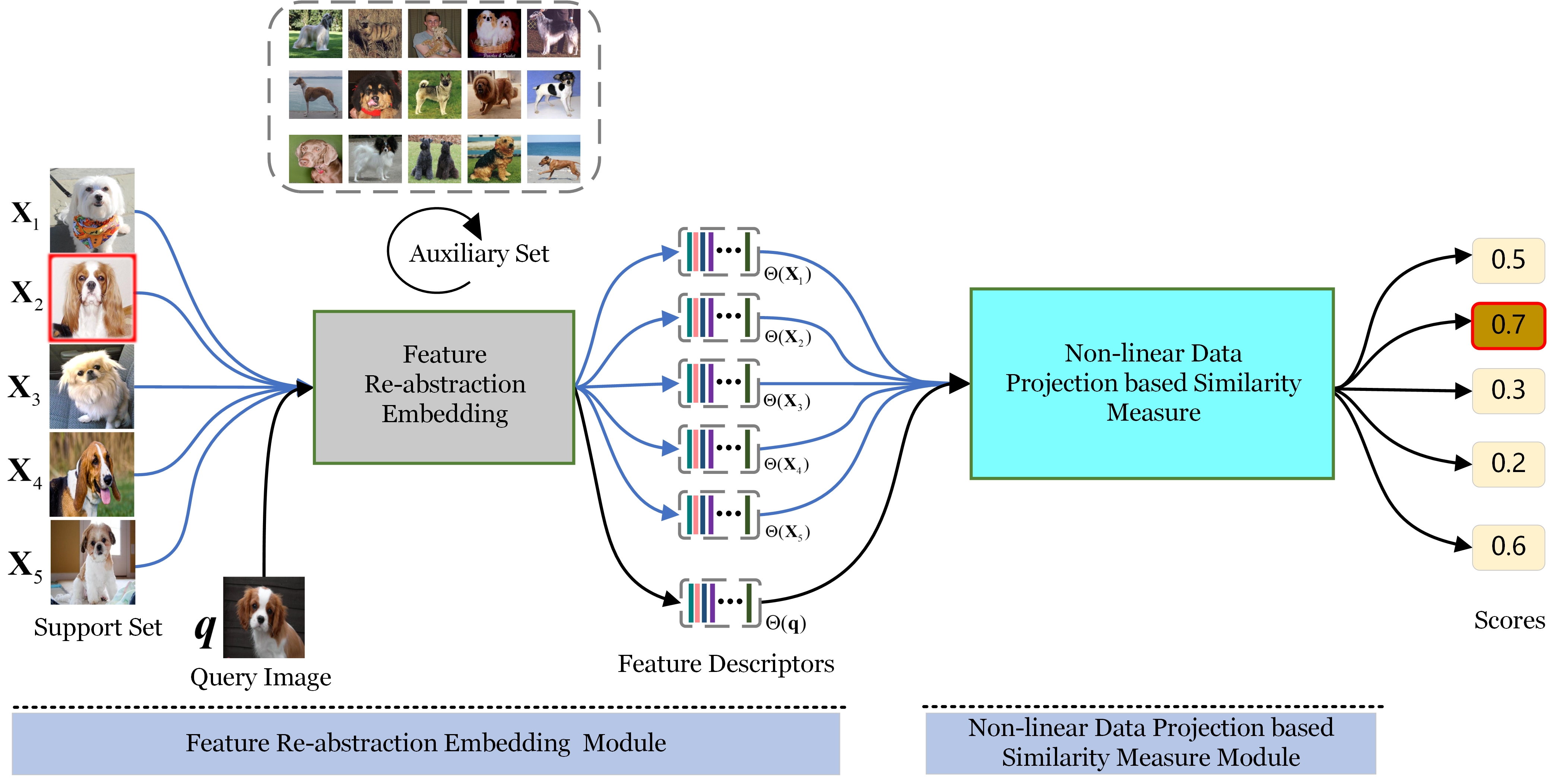} % Reduce the figure size so that it is slightly narrower than the column. Don't use precise values for figure width.This setup will avoid overfull boxes.
\caption{The framework of the proposed NDPNet network for a 5-way 1-shot FSFGIC task which contains two modules. (1) A feature re-abstraction embedding module. (2) A non-linear data projection based similarity measure module.}
\label{fig1}
\end{figure*}

The aforementioned metric-learning based methods utilized image level global features to represent each sample and perform few-shot learning tasks. It is worth to note that the training examples of each class are extremely limited in few-shot learning. Because local feature descriptors can more effectively represent the distribution of each category than image level global features in few-shot learning~\cite{li2019distribution,ijcai2020-100,li2020revisiting}, some recent work (e.g., covariance metric networks (CovaMNet)~\cite{li2019distribution}, adaptive task-aware local representations network (ATL-Net)~\cite{ijcai2020-100}, and deep nearest neighbor neural network (DN4)~\cite{li2020revisiting}) employed the local feature descriptors to represent each sample. And then, covariance metric~\cite{li2019distribution}, adaptive threshold for episodic attention based on classification probability~\cite{ijcai2020-100}, and cosine metric~\cite{li2020revisiting} were utilized for measuring the relationship between a query sample and each support class for FSFGIC. Compared with the aforementioned metric-learning methods (e,g., the CovaMNet~\cite{li2019distribution}, ATL-Net~\cite{ijcai2020-100}, and DN4~\cite{li2020revisiting}), the feature descriptors extracted by the proposed feature re-abstraction embedding network designed in this work has re-enhanced finer details into abstract feature representation. More important, projecting samples from query images and samples from support sets into a non-linear space with different projection factors and letting the network learn appropriate projection factors can make the distance between samples within a category smaller and the distance between samples from different categories larger, and thus can effectively improve the classification performance in FSFGIC. Experiments on fine-gained image classification benchmark datasets exhibit the superior performance from our proposed method compared with metric learning benchmarks state-of-the-arts.

\section{Methodology}
\subsection{Problem Statement}
A common few-shot fine-grained dataset includes two parts: a support set $\mathcal{\boldsymbol{S}}$ and a query set $\mathcal{\boldsymbol{Q}}$. The small support set $\mathcal{\boldsymbol{S}}$ contains $\mathcal{C}$ unseen classes, each of which has $\mathcal{K}$ labeled samples. The query set $\mathcal{\boldsymbol{Q}}$ contains unlabeled samples. Sets $\mathcal{\boldsymbol{S}}$ and $\mathcal{\boldsymbol{Q}}$ share the same label space. The goal of FSFGIC is to successfully classify each query sample $\boldsymbol{q}$ ($\boldsymbol{q}\in\mathcal{\boldsymbol{Q}}$) into a corresponding class in $\mathcal{C}$. Thus, the problem is noted as a $\mathcal{C}$-way $\mathcal{K}$-shot task. However, the training samples of each class are too limited to effectively learn transferable knowledge~\cite{ijcai2020-100}.

In this work, an auxiliary set $\mathcal{\boldsymbol{A}}$ and an episodic training paradigm~\cite{vinyals2016matching} are utilized for tackling the aforementioned problem. The auxiliary set $\mathcal{\boldsymbol{A}}$ consists of a large quantity of classes and labeled samples which are far larger than $\mathcal{C}$ and $\mathcal{K}$ respectively and can be divided into many $\mathcal{C}$-way $\mathcal{K}$-shot FSFGIC tasks $\mho$. It is worth to note that there is no intersection between the label space sets $\mathcal{\boldsymbol{A}}$ and $\mathcal{\boldsymbol{S}}$. For each $\mho$, it consists of an auxiliary support set $\mathcal{\boldsymbol{A_S}}$ and an auxiliary query set $\mathcal{\boldsymbol{A_Q}}$. In the episodic training stage, thousands of tasks $\mho$ will be constructed for training the transferable knowledge. Once the transferable knowledge is obtained, each sample from $\mathcal{\boldsymbol{Q}}$ will be classified into one class in set $\mathcal{\boldsymbol{S}}$.

\subsection{The Proposed NDPNet}
The overview of our proposed NDPNet framework for FSFGIC is illustrated in Fig.~\ref{fig1}. It contains two main modules: a feature re-abstraction embedding (FRaE) module and a non-linear data projection based similarity measure (NDP-SM) module.~The feature re-abstraction embedding module is designed to feature representations with detail re-enhanced semantic information for all samples. After that, a similarity measure module is designed to compute the relationships of the learned feature representations between a query sample $\mathcal{\boldsymbol{q}}$ and each support class $\mathcal{\boldsymbol{S}}$. Predominantly, our proposed NDPNet architecture can be trained in an end-to-end manner from scratch. In the following, the proposed feature re-abstraction embedding module and the similarity measure module are presented in detail.

\begin{figure}[!htbp]
\setlength{\abovecaptionskip}{5pt}
\setlength{\belowcaptionskip}{5pt}
\centering
\includegraphics[width=1.01\columnwidth]{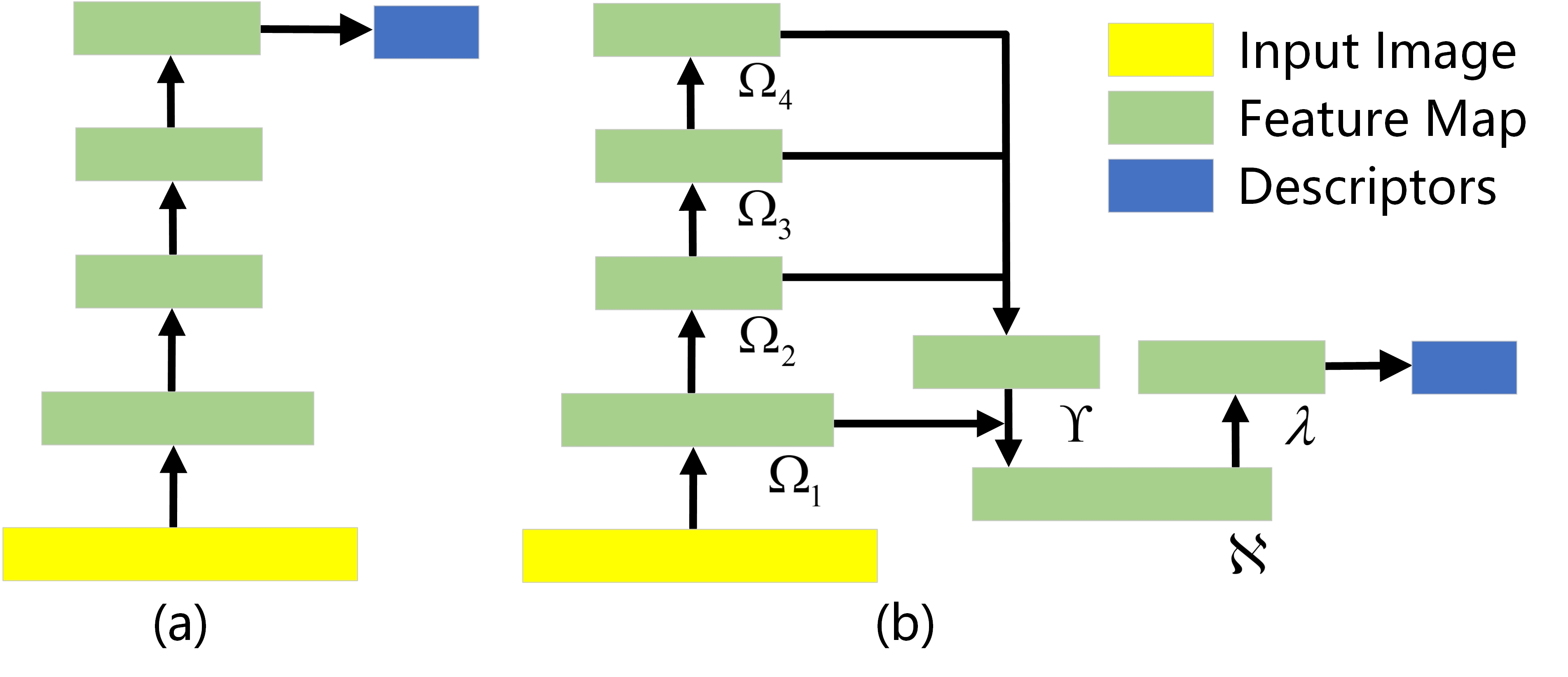} % Reduce the figure size so that it is slightly narrower than the column. Don't use precise values for figure width.This setup will avoid overfull boxes.
\caption{(a) The framework of the existing four convolutional blocks. (b) Our proposed feature re-abstraction embedding network.}
\label{fig4}
\end{figure}

{\bf Feature Re-abstraction Embedding Module.} Four convolutional blocks (i.e.,~\textit{Conv-64F}~\cite{vinyals2016matching}) are widely used in meta-learning based FSFGIC tasks~\cite{finn2017model, munkhdalai2018rapid, zhu2020multi} and metric-learning based FSFGIC tasks~\cite{vinyals2016matching, snell2017prototypical, satorras2018few, li2019distribution, li2020revisiting, huang2020low, ijcai2020-100}. They utilized the most abstract descriptors from the deepest layer that contains strong semantic information for FSFGIC as shown in Fig.~\ref{fig4}(a). Here, we design a novel feature re-abstraction embedding network as shown in Fig.~\ref{fig4}(b), to re-enhance the most abstract feature with finer details ($\Omega_{1}$, $\Omega_{2}$, $\Omega_{3}$, $\Omega_{4}$) and re-abstract back to its most abstract form for metric learning in the next module.

%However, they ignore the impact of image detail information on classification performance. In order to extract descriptors with SSDE from input images with a single scale, we design a novel feature re-abstraction embedding network as shown in Fig.~\ref{fig4}(b). It can be seen from Fig.~\ref{fig4}(b) that our designed feature re-abstraction embedding network combines strong semantic information and fine image detail information to make descriptors have strong semantic and detail enhancement characteristics.

%It is well known that we need to use multiple scale techniques~\cite{witkin1987scale, zhang2020corner} to describe a local object in the image for knowing the best scale required to describe the object. Meanwhile, the size of the same object in the image may vary with different image scale transformations~\cite{zhang2019corner2, lindeberg1993detecting}. Therefore, it is necessary for us to consider how to design a scale invariant embedding network to obtain local feature information with SI for performing FSFGIC tasks.

\begin{figure*}[!htbp]
\setlength{\abovecaptionskip}{5pt}
\setlength{\belowcaptionskip}{5pt}
\centering
\includegraphics[width=2.07\columnwidth]{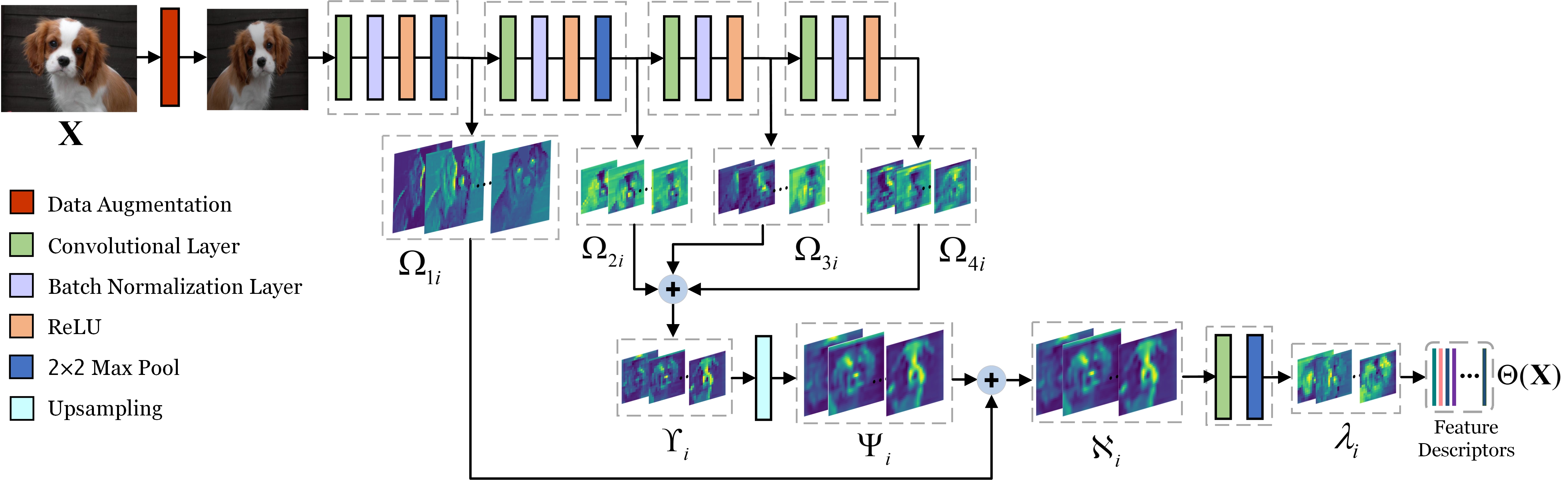} % Reduce the figure size so that it is slightly narrower than the column. Don't use precise values for figure width.This setup will avoid overfull boxes.
\caption{The feature re-abstraction embedding module.}
\label{fig2}
\end{figure*}

The detailed framework of our designed feature re-abstraction embedding network is shown in Fig.~\ref{fig2}. Each convolutional block consists of a convolutional layer, a batch normalization layer, and a Leaky ReLU layer. In addition, a $2\times 2$ max pooling layer is attached to the first two convolutional blocks. It is worth to note that 64 filters with a size of $3\times 3$ are utilized in the convolutional layer. For an input image $\mathit{\boldsymbol{X}}$ ($\mathit{\boldsymbol{X}}\in\mathcal{\boldsymbol{S}}\cup\mathcal{\boldsymbol{Q}}$) with a size of $N\times N$, through the first convolutional block, 64 feature tensors with a size of $\frac{N}{2}\times \frac{N}{2}$ can be obtained which is named as $\Omega_{1i}$ ($i=1,2,...,64$). And then three sets of 64 feature tensors with a size of $\frac{N}{4}\times \frac{N}{4}$, which are named as $\Omega_{2i}$, $\Omega_{3i}$, and $\Omega_{4i}$ ($i=1,2,...,64$), can be obtained after the second, third, and fourth convolutional blocks.

%Inspired by the FPN method~\cite{lin2017feature}, low resolution with strong semantic features and the high resolution with weak semantic features are combined through a top-down path and horizontal connections for obtaining a feature pyramid with strong semantics at different scales. In this work, local feature tensors $\Omega_{2i}$, $\Omega_{3i}$, and $\Omega_{4i}$ ($i=1,2,...,64$) are combined as follows

In this work, feature tensors obtained from the second, third, and fourth convolution block (i.e., $\Omega_{2i}$, $\Omega_{3i}$, and $\Omega_{4i}$ ($i=1,2,...,64$)) are combined as follows
\begin{equation}\begin{aligned}
\label{eq1}
\Upsilon_{i}&=\Omega_{2i}+\Omega_{3i}+\Omega_{4i},\\
i&=1,2,...,64.
\end{aligned}\end{equation}
For the combined tensors $\Upsilon_{i}$ ($i=1,2,...,64$), we upsample their corresponding spatial resolution with a scaling factor of 2
\begin{equation}\begin{aligned}
\label{eq2}
\Psi_{i}&=\uparrow_{2}\Upsilon_{i},\\
i&=1,2,...,64,
\end{aligned}\end{equation}
where $\uparrow_{2}$ denotes a bilinear upsampling operator with the scaling factor of 2. And then, the upsampled tensor $\Psi_{i}$ is merged with tensor $\Omega_{1i}$ ($i=1,2,...,64$) as
\begin{equation}\begin{aligned}
\label{eq3}
\aleph_{i}&=\Psi_{i}+\Omega_{1i},\\
i&=1,2,...,64.
\end{aligned}\end{equation}
Furthermore, a convolutional layer (with 64 filters of size 3$\times$3) is attached to the merged tensor $\aleph_{i}$ for reducing the aliasing effect caused by upsampling and obtaining 64 tensors $\xi_{i}$ ($i=1,2,...,64$) with a size of $\frac{N}{2}\times \frac{N}{2}$. Then, a $2\times 2$ max pooling layer is attached to the convolutional layer for obtaining feature tensor maps $\lambda_{i}$ ($i=1,2,...,64$). Finally, the feature values at the same coordinate position on different feature tensor maps are used to construct a feature descriptor. The descriptor is constructed from feature maps of different resolutions, and it contains the semantic information of the original deepest feature tensor map, so it has strong detail re-enhanced semantic characteristics.

%For a test image, $\frac{N}{4}\times \frac{N}{4}$ local feature descriptors are constructed.

In conclusion, through the designed feature embedding module, an input image $\mathit{\boldsymbol{X}}$ can be represented as a group of $d\times(h\times w)$-dimensional descriptors as follows
\begin{equation}\begin{aligned}
\label{eq4}
\Theta(\mathit{\boldsymbol{X}})=[\boldsymbol{x}_{1},...,\boldsymbol{x}_{j},...,\boldsymbol{x}_{m}]\in R^{d\times(h\times w)},
\end{aligned}\end{equation}
where $\boldsymbol{x}_{j}$ is the $j$-th descriptor, $h$ and $w$ denote the height and width of the feature tensor map respectively, $d$ (here $d=64$) is the number of filters, $R$ represents real space, and $m$ ($m=h\times w$) is the total number of descriptors for the input image $\mathit{\boldsymbol{X}}$. Meanwhile, the $m$ descriptors can be used as the representations of image $\mathit{\boldsymbol{X}}$.

\begin{figure*}[!htbp]
\setlength{\abovecaptionskip}{5pt}
\setlength{\belowcaptionskip}{5pt}
\centering
\includegraphics[width=2.05\columnwidth]{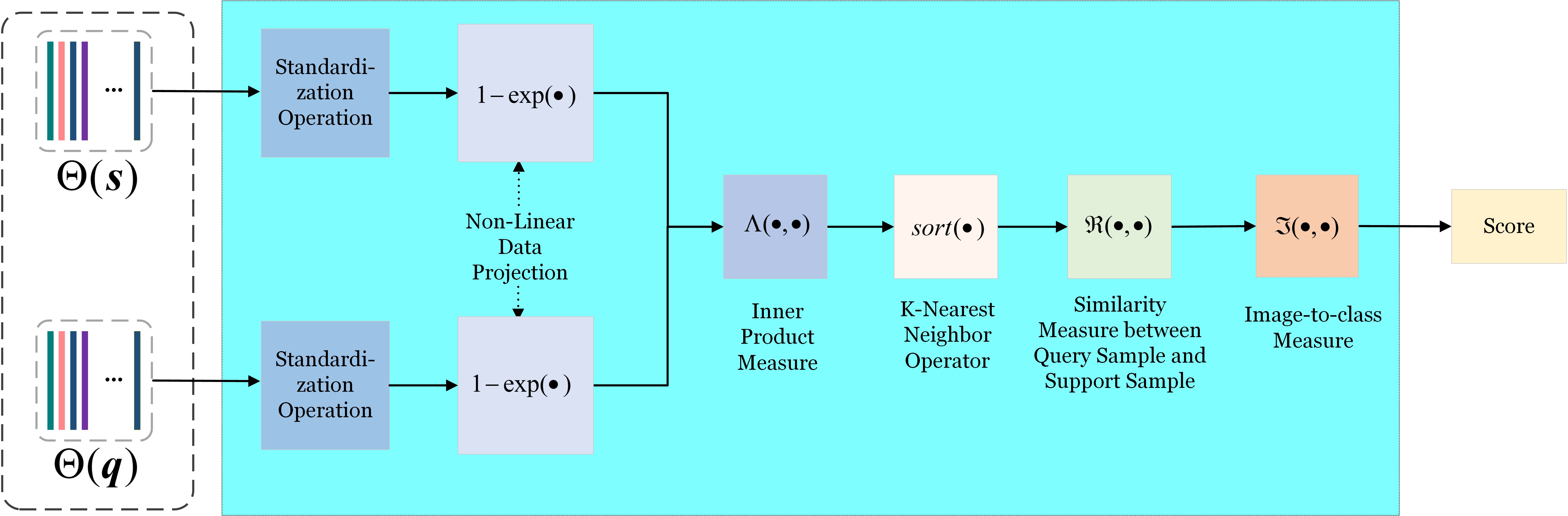} % Reduce the figure size so that it is slightly narrower than the column. Don't use precise values for figure width.This setup will avoid overfull boxes.
\caption{The framework of non-linear data projection based similarity measure module.}
\label{fig3}
\end{figure*}

{\bf Non-linear Data Projection based Metric Learning.} In this module, a novel non-linear data projection strategy is designed to make the distance between the descriptors within a class smaller and the distance between descriptors of different classes larger.

Through the designed feature re-abstraction embedding module, the descriptors of a query sample $\mathit{\boldsymbol{q}}$ and a support sample $\mathit{\boldsymbol{s}}$ can be represented as $\Theta(\mathit{\boldsymbol{q}})=[{\boldsymbol{q}}_{1},...,{\boldsymbol{q}}_{m}]\in R^{d\times(h\times w)}$ and $\Theta(\mathit{\boldsymbol{s}})=[{\boldsymbol{s}}_{1},...,{\boldsymbol{s}}_{m}]\in R^{d\times(h\times w)}$ respectively. For descriptors ${\boldsymbol{q}}_{j}$ and ${\boldsymbol{s}}_{j}$ $(j=1,...,m)$, standardization operation is performed as
\begin{equation}\begin{aligned}
\label{eq99}
\ddot{\boldsymbol{q}}_{j}&=\frac{\boldsymbol{q}_j-M(\boldsymbol{q}_j)}{\sqrt{D(\boldsymbol{q}_j)}},\\
\ddot{\boldsymbol{s}}_{j}&=\frac{\boldsymbol{s}_j-M(\boldsymbol{s}_j)}{\sqrt{D(\boldsymbol{s}_j)}},\\
j&=1,...,m,
\end{aligned}\end{equation}
where $M(\cdot)$ denotes the mean measure and $D(\cdot)$ denotes the variance measure. For each element $a_{u}$ and $b_{u}$ $(u=1,...,64)$ in the standardized descriptors $\ddot{\boldsymbol{q}}_{j}$ and $\ddot{\boldsymbol{s}}_{j}$~$(j=1,...,m)$, a non-linear projection is designed as follows
\begin{equation}\begin{aligned}
\label{eq9}
\hat{a}_{u}&=\text{sign}(a_{u})\bigg(1-\text{exp}\bigg(-\frac{a^2_{u}}{2\alpha^2}\bigg)\bigg),\\
\hat{b}_{u}&=\text{sign}(b_{u})\bigg(1-\text{exp}\bigg(-\frac{b^2_{u}}{2\beta^2}\bigg)\bigg),
\end{aligned}\end{equation}
where $\alpha$ and $\beta$ denote the non-linear projection factors and $\text{sign}(\cdot)$ denotes the $\text{sign}$ function. Then the descriptors of a query sample $\mathit{\boldsymbol{q}}$ after non-linear projection and a support sample $\mathit{\boldsymbol{s}}$ after non-linear projection can be expressed as $\Theta(\mathit{\boldsymbol{q}})=[\hat{\boldsymbol{q}}_{1},...,\hat{\boldsymbol{q}}_{m}]\in R^{d\times(h\times w)}$ and $\Theta(\mathit{\boldsymbol{s}})=[\hat{\boldsymbol{s}}_{1},...,\hat{\boldsymbol{s}}_{m}]\in R^{d\times(h\times w)}$ respectively.

The framework of non-linear data projection based similarly metric-learning module is shown in Fig.~\ref{fig3}. In this work, the inner product is employed for measuring the similarities of feature information between a query image after non-linear projection and a support class after non-linear projection for performing FSFGIC tasks.

%and $\hat{\boldsymbol{s}}_{j}$

For each $\hat{\boldsymbol{q}}_{j}$~$(j=1,...,m)$, the inner product measure is used to find its corresponding $k$-nearest neighbors $\boldsymbol{\alpha}_{j}^{t}\vert_{t=1}^{k}~(j=1,...,m)$ from $\Theta(\mathit{\boldsymbol{s}})$ based on the nearest-neighbor method~\cite{boiman2008defense} as follows
\begin{equation}\begin{aligned}
\label{eq5}
&\Lambda(\hat{\boldsymbol{q}}_{j},\hat{\boldsymbol{s}}_{\tau})=\vert<\hat{\boldsymbol{q}}_{j},\hat{\boldsymbol{s}}_{\tau}>\vert,\\
&\zeta=sort\{\Lambda(\hat{\boldsymbol{q}}_{j},\hat{\boldsymbol{s}}_{1}),...,\Lambda(\hat{\boldsymbol{q}}_{j},\hat{\boldsymbol{s}}_{m})\},\\
&\boldsymbol{\alpha}_{j}^{t}=\hat{\boldsymbol{s}}_{\zeta_{t}},\\
&t=1,...,k,~j=1,...,m,~\tau=1,...,m,
\end{aligned}\end{equation}
where $<\cdot>$ stands for an inner product measure, $\vert\cdot\vert$ stands for an absolute value function, $sort$ denotes the operation of arranging the values of the array from the largest to the smallest, $\zeta$ represents the sequence number corresponding to the descriptor $\hat{\boldsymbol{s}}_{\tau}$ ($\tau=1,...,m$) arranged in descending order of the similarity with the descriptor $\hat{\boldsymbol{q}}_{j}$, $\boldsymbol{\alpha}_{j}^{t}$ represents $k$ descriptors $\hat{\boldsymbol{s}}_{\zeta_{t}}~(t=1,...,k)$ that are most similar to descriptor $\hat{\boldsymbol{q}}_{j}$. Then, the non-linear data projection based similarity measure $\Re(\mathit{\boldsymbol{q}},\mathit{\boldsymbol{s}})$ between a query sample $\mathit{\boldsymbol{q}}$ and a support sample $\mathit{\boldsymbol{s}}$ is defined as
\begin{equation}\begin{aligned}
\label{eq6}
\Re(\mathit{\boldsymbol{q}},\mathit{\boldsymbol{s}})=\sum_{j=1}^{m}\sum_{t=1}^{k}\vert<\hat{\boldsymbol{q}}_{j},\boldsymbol{\alpha}_{j}^{t}>\vert.
%Cov(\hat{q}_{j},\hat{s}_{j}^{v})
\end{aligned}\end{equation}
In this work, $k$ is set to 1.

Furthermore, we extend the non-linear projection based inner product similarity measure to compute the similarity between a query sample and each support class for FSFGIC tasks. Specifically, the descriptors of a query sample $\mathit{\boldsymbol{q}}$ and a support class $\boldsymbol{\mathit{S}}$ can be represented as $\Theta(\mathit{\boldsymbol{q}})=[\hat{\boldsymbol{q}}_{1},...,\hat{\boldsymbol{q}}_{m}]\in R^{d\times(h\times w)}$ and $\Theta(\boldsymbol{\mathit{S}})=[\Theta(\mathit{\boldsymbol{X}_1}),...,\Theta(\mathit{\boldsymbol{X}_{\mathcal{K}}})]\in R^{d\times(h\times w\times {\mathcal{K}})}$ respectively, where ${\mathcal{K}}$ is the number of samples in support class $\boldsymbol{\mathit{S}}$. From Equation~(\ref{eq5}), each $\hat{\boldsymbol{q}}_{j}~(j=1,...,m)$'s corresponding $1$-nearest neighbor from the support class $\boldsymbol{\mathit{S}}$ can be represented as $\hat{\boldsymbol{s}}_{j}~(j=1,...,m)$. The non-linear data projection based image-to-class similarity measure~\cite{li2020revisiting}) is defined as
\begin{equation}\begin{aligned}
\label{eq7}
\Im(\mathit{\boldsymbol{q}},\boldsymbol{\mathit{S}})=\sum_{j=1}^{m}\vert<\hat{\boldsymbol{q}}_{j},\hat{\boldsymbol{s}}_{j}>\vert.
%Cov(\hat{q}_{j},\hat{s}_{j}^{v})
\end{aligned}\end{equation}

And then the Adam optimization method~\cite{kingma2014adam} with a cross-entropy loss is used to train the whole network for learning the weights (including the appropriate non-linear projection factors) and performing FSFGIC tasks.

\section{Experiments}
\subsection{Datasets}
Our proposed network is evaluated on four fine-gained datasets of Stanford Dogs~\cite{khosla2011novel}, Stanford Cars~\cite{krause20133d}, CUB-200~\cite{welinder2010caltech}, and Cottons~\cite{yu2020patchy}.~The Stanford Dogs dataset consists of 120 dog classes with 20,580 samples.~The Stanford Cars dataset consists of 196 car classes with 16,185 samples. The CUB-200 dataset consists of 200 bird classes with 6,033 samples. The Cottons dataset consists of 80 cotton classes with 480 samples. For fair performance comparisons, we follow the same data split as used in~\cite{ijcai2020-100} that are illustrated in Table~\ref{tab1}.

\begin{table}[htbp]

\centering
\begin{tabular}{cccc}
\hline
{Dataset}  &$N_{train}$   & $N_{val}$  & $N_{test}$   \\
\hline
Stanford Dogs      &70  & 20 & 30    \\
Stanford Cars        & 130  & 17  & 49        \\
CUB-200      & 130  & 20  & 50         \\
Cottons      & 51  & 13  & 16         \\
\hline
\end{tabular}
\setlength{\abovecaptionskip}{10pt}
\setlength{\belowcaptionskip}{10pt}
\caption{The class split of four fine-grained datasets. $N_{train}$, $N_{val}$, and $N_{test}$ are the numbers of classes in the auxiliary set, validation set, and test set respectively.}
\label{tab1}
\end{table}

%\begin{table}[htbp]
%\centering
%\begin{tabular}{cccc}
%\hline
%{Dataset}  &$N_{train}$   & $N_{val}$  & $N_{test}$   \\
%\hline
%Stanford Dogs      &70  & 20 & 30    \\
%Stanford Cars        & 130  & 17  & 49        \\
%CUB-200      & 130  & 20  & 50         \\
%Cottons      & 51  & 13  & 16         \\
%\hline
%\end{tabular}
%\caption{The class split of four fine-grained datasets. $N_{train}$, $N_{val}$, and $N_{test}$ are the number of classes in the auxiliary set, validation set, and test set respectively.}
%\label{tab1}
%\end{table}

\subsection{Experimental Setup}
In this work, both the 5-way 1-shot and 5-way 5-shot FSFGIC tasks are performed on the aforementioned four datasets. Each input image is resized to a fixed size of $100\times100$ and randomly cropped into $84\times84$. Random affine transformations, random horizontal flips, and random rotations are utilized for data augmentation. We randomly sample and construct 300,000 episodes for training our models by utilizing the episodic training paradigm~\cite{vinyals2016matching}. For each episode, 15 query samples per class are randomly selected for the datasets of Stanford Dogs, Stanford Cars, and CUB-200 respectively. Because each class has only six samples in the Cottons dataset, 5 and 1 query samples are selected from each cotton class for the 1-shot and 5-shot settings respectively. The Adam optimization method~\cite{kingma2014adam} is utilized for training the models using 30 epochs. The learning rate is initially set as 0.001 and multiplied by 0.5 for every 100,000 episodes.

In the testing stage, 600 episodes are randomly constructed from the testing set for obtaining the results. The top-1 mean accuracy is employed as the evaluation criteria. The aforementioned process is repeated five times and the final mean results are obtained as the classification accuracy of FSFGIC. Meanwhile, the $95\%$ confidence intervals are obtained and reported.

\begin{table*}[htp]

\renewcommand{\arraystretch}{1.2}
\centering

\begin{tabular}{ccccccc}

\hline
%\multirow{Model} & \multirow{Base Model} & \multicolumn{6}{c}{Accuracy (\%)}\\
\multirow{3}*{Model}  & \multicolumn{5}{c}{5- Way Accuracy (\%)}\\
\cline{2-7}
 & \multicolumn{2}{c}{Stanford Dogs} &\multicolumn{2}{c}{Stanford Cars} &\multicolumn{2}{c}{CUB-200}\\
 \cline{2-3} \cline{4-5}  \cline{6-7}
  & {5-way 1-shot} &{5-way 5-shot} &{5-way 1-shot} & {5-way 5-shot}&{5-way 1-shot} &{5-way 5-shot} \\
 \hline
 Matching Net        &35.80$\pm$0.99 &47.50$\pm$1.03 &34.80$\pm$0.98 &44.70$\pm$1.03 &45.30$\pm$1.03 &59.50$\pm$1.01\\
 P-Net    &37.59$\pm$1.00 &48.19$\pm$1.03 &40.90$\pm$1.01 &52.93$\pm$1.03 &37.36$\pm$1.00 &45.28$\pm$1.03\\
 GNN                 &46.98$\pm$0.98 &62.27$\pm$0.95 &55.85$\pm$0.97 &71.25$\pm$0.89 &51.83$\pm$0.98 &63.69$\pm$0.94\\
 CovaMNet            &49.10$\pm$0.76 &63.04$\pm$0.65 &56.65$\pm$0.86 &71.33$\pm$0.62 &52.42$\pm$0.76 &63.76$\pm$0.64\\
 DN4                 &45.41$\pm$0.76 &63.51$\pm$0.62 &59.84$\pm$0.80 &88.65$\pm$0.44 &46.84$\pm$0.81 &74.92$\pm$0.64\\
PABN$+_{cpt}$              &45.65$\pm$0.71 &61.24$\pm$0.62 &54.44$\pm$0.71 &67.36$\pm$0.61 &63.36$\pm$0.80 &74.71$\pm$0.60\\
$\text{LRPABN}_{cpt}$  &45.72$\pm$0.75 &60.94$\pm$0.66 &60.28$\pm$0.76 &73.29$\pm$0.58 &63.63$\pm$0.77 &76.06$\pm$0.58\\
 ATL-Net             &54.49$\pm$0.92 &73.20$\pm$0.69 &67.95$\pm$0.84 &89.16$\pm$0.48 &60.91$\pm$0.91 &77.05$\pm$0.67\\

Our NDPNet           &{\bf 56.21$\pm$0.86} & {\bf 74.82$\pm$0.84} & {\bf 71.48$\pm$0.89 }&{\bf  91.92$\pm$0.91 }&  {\bf64.74$\pm$0.90} &{\bf 80.52$\pm$0.63}\\

\hline
%Ours & VGG-16 & 53.24&46.60&84.20&91.06&88.52&96.62\\
%Ours & ResNet-50&\textbf{60.83}&\textbf{53.67}&\textbf{85.78}&\textbf{92.29}&\textbf{90.88}&\textbf{97.16} \\

%\hline
\end{tabular}
\caption{\label{tab:test}Comparison results on three different standard datasets.}
\label{t2}
\end{table*}

\subsection{Performance Comparison}
The experimental results on Stanford Dogs, Stanford Cars, CUB-200, and Cottons datasets are presented in Table~\ref{t2} and Table~\ref{t3} and compared with eight state-of-the-art metric learning methods (i.e., Matching Net~\cite{vinyals2016matching}, Prototypical Net (P-Net)~\cite{snell2017prototypical}, GNN~\cite{satorras2018few}, CovaMNet~\cite{li2019distribution}, DN4~\cite{li2020revisiting}, PABN$+_{cpt}$~\cite{huang2020low}, $\text{LRPABN}_{cpt}$~\cite{huang2020low}, and ATL-Net~\cite{ijcai2020-100}). From Table~\ref{t2} and Table~\ref{t3}, it is observed that our proposed NDPNet method outperforms all the benchmark metric-learning methods on both 5-way 1-shot and 5-way 5-shot FSFGIC tasks. For the 5-way 1-shot and 5-way 5-shot FSFGIC tasks on the Stanford Cars dataset, our proposed NDPNet architecture achieves $36.68\%$, $30.58\%$, $15.63\%$, $14.83\%$, $11.64\%$, $17.04\%$, $11.20\%$, and $3.53\%$ improvements and $47.22\%$, $38.99\%$, $20.67\%$, $20.59\%$, $3.27\%$, $24.56\%$, $18.63\%$, and $2.76\%$ improvements over Matching Net, P-Net, GNN, CovaMNet, DN4, PABN$+_{cpt}$, $\text{LRPABN}_{cpt}$, and ATL-Net respectively. Such improvements is due to the ability of NDPNet for making the distance between the samples within the same class smaller and the distance between samples from different classes larger and reducing the coupling relationship between samples of different categories.

%that the designed non-linear data projection based similarly metric learning network has the ability to make the distance between the samples within the same class smaller and the distance between samples from different classes larger and reduce the coupling relationship between samples of different categories. Furthermore, the proposed NDPNet architecture has considered how to obtain local feature representations with SSDE.

\begin{table}[htp]
\renewcommand{\arraystretch}{1.2}
\centering

\begin{tabular}{ccc}

\hline
\multirow{2}*{Model}   & \multicolumn{2}{c}{5-Way Accuracy (\%)}\\
\cline{2-3}
  & {1-shot} &{5-shot} \\
 \hline
 Matching Net            & 38.77$\pm$1.08 & 58.95$\pm$1.02\\
 P-Net                   & 39.12$\pm$1.03 & 60.25$\pm$1.02\\
 GNN                    & 43.91$\pm$0.71 & 68.61$\pm$0.81\\
 CovaMNet                & 46.14$\pm$0.91 & 69.97$\pm$0.89\\
  DN4                    & 45.74$\pm$0.86 & 70.36$\pm$0.95\\
PABN$+_{cpt}$            & 46.73$\pm$0.77 & 70.32$\pm$0.71\\
 $\text{LRPABN}_{cpt}$   & 46.81$\pm$0.73 & 71.82$\pm$0.99\\
 ATL-Net                 & 51.89$\pm$0.97 & 73.48$\pm$0.96\\

Our NDPNet            & {\bf 53.52$\pm$0.98} & {\bf 74.66$\pm$0.92}\\

\hline

\end{tabular}
\caption{\label{tab:test}Comparison results on the Cottons dataset.}
\label{t3}
\end{table}

\subsection{Discussion}
Below we conduct several further experiments for studying the effectiveness of the presented NDPNet architecture.

{\bf Influence of the feature re-abstraction embedding network for FSFGIC.} In this experiment, the designed embedding network in our architecture is replaced with the basic embedding network (i.e., \textit{Conv-64F}). Here, \textit{Conv-64F} is combined with our non-linear data projection based similarity measure ($\text{NDP-SM}$) module, which is denoted as $\textit{Conv-64F}+\text{NDP-SM-module}$ for performing FSFGIC tasks on the Stanford Dogs and Stanford Cars datasets. Their corresponding results on the 5-way 1-shot and 5-way 5-shot tasks are presented in Table~\ref{t4}. We can see from Table~\ref{t4} that the classification accuracies of our proposed $\text{NDPNet}$ is far better than the classification accuracy of the $\textit{Conv-64F}+\text{NDP-SM-module}$, which demonstrates the effectiveness of the proposed feature re-abstraction embedding (FRaE) module in image feature description for metric learning.

% The reason is that the presented NDPNet architecture has considered how to effectively obtain feature information with strong semantic and detail enhancement from images.

\begin{table*}[htp]
\renewcommand{\arraystretch}{1.2}
\centering

\begin{tabular}{ccccc}

\hline
%\multirow{Model} & \multirow{Base Model} & \multicolumn{6}{c}{Accuracy (\%)}\\
\multirow{3}*{Model}  & \multicolumn{4}{c}{5- Way Accuracy (\%)}\\
\cline{2-5}
 & \multicolumn{2}{c}{Stanford Dogs} &\multicolumn{2}{c}{Stanford Cars} \\
 \cline{2-3} \cline{4-5}
  & {5-way 1-shot} &{5-way 5-shot} &{5-way 1-shot} & {5-way 5-shot} \\
 \hline
$\textit{Conv-64F}+\text{NDP-SM-module}$      & 53.17$\pm$0.92 & 69.26$\pm$0.98 &67.80$\pm$0.99 &89.70$\pm$0.83 \\
$\text{FRaENet}$               &52.59$\pm$1.03 &68.37$\pm$1.12 &65.97$\pm$0.98 &86.93$\pm$1.07\\
$\text{NDPNet}\_{\textit{Cos}}$        &55.38$\pm$0.96 &74.71$\pm$0.95 &71.10$\pm$0.98 &91.55$\pm$0.89 \\
$\text{GK\_FRaENet}$               &54.28$\pm$0.85 &73.57$\pm$0.65 &69.15$\pm$0.92 &89.25$\pm$0.91 \\
Our $\text{NDPNet}$           &{\bf 56.21$\pm$0.86} & {\bf 74.82$\pm$0.84} & {\bf 71.48$\pm$0.89 }&{\bf  91.92$\pm$0.91 }\\

\hline

\end{tabular}
\caption{\label{tab:test}The results of ablation study on the proposed NDPNet on the Stanford Dogs and Stanford Cars datasets.}
\label{t4}
\end{table*}

\begin{table*}[htp]
\renewcommand{\arraystretch}{1.2}
\centering
\begin{tabular}{ccccccc}

\hline
%\multirow{Model} & \multirow{Base Model} & \multicolumn{6}{c}{Accuracy (\%)}\\
\multirow{3}*{Model}  & \multicolumn{5}{c}{5- Way Accuracy (\%)}\\
\cline{2-7}
 & \multicolumn{2}{c}{Stanford Dogs} &\multicolumn{2}{c}{Stanford Cars} &\multicolumn{2}{c}{CUB-200}\\
 \cline{2-3} \cline{4-5}  \cline{6-7}
  & {5-way 1-shot} &{5-way 5-shot} &{5-way 1-shot} & {5-way 5-shot}&{5-way 1-shot} &{5-way 5-shot} \\
 \hline

 $\text{NDPNet}\_\{k=5\}$          &54.71$\pm$0.74 &73.33$\pm$0.91  &70.11$\pm$0.77 &89.52$\pm$0.83 &63.18$\pm$0.88 &79.73$\pm$0.79\\
$\text{NDPNet}\_\{k=3\}$           &55.78$\pm$0.89 &73.91$\pm$0.97  &70.75$\pm$0.92 &90.31$\pm$0.99 &64.55$\pm$0.98 &80.23$\pm$0.81 \\
 $\text{NDPNet}\_\{k=1\}$           &{\bf 56.21$\pm$0.86} & {\bf 74.82$\pm$0.84} & {\bf 71.48$\pm$0.89 }&{\bf  91.92$\pm$0.91 }&  {\bf64.74$\pm$0.90} &{\bf 80.52$\pm$0.63}\\

\hline

\end{tabular}
\caption{\label{tab:test}The impact of the $k$-nearest neighbors on the proposed NDPNet on three different standard datasets.}
\label{t5}
\end{table*}

{\bf Influence of the non-linear data projection for FSFGIC.} In the following experiments, we firstly consider the classification performance of a network after removing the designed non-linear data projection strategy, which is named as $\text{FRaENet}$. The descriptors from query images and the descriptors from support classes obtained from the designed embedding network are utilized to calculate the similarities using Equations~(\ref{eq5}),~(\ref{eq6}), and~(\ref{eq7}). The $\text{FRaENet}$ is employed for performing FSFGIC tasks on the Stanford Dogs and Stanford Cars datasets. Their results on the 5-way 1-shot and 5-way 5-shot tasks are presented in Table~\ref{t4}. After removing the proposed non-linear data projection based similarity measure ($\text{NPD-SM}$) module, the classification accuracy of $\text{FRaENet}$ is significantly lower than the proposed NDPNet, which validates the effectiveness and strength of the proposed $\text{NPD-SM}$ network.

%greatly reduced. The reason is that $\text{FRENet}$ module cannot effectively reduce the coupling relationship between samples from different categories.

Secondly, the inner product similarity measure in our architecture is replaced with the cosine similarity measure, which is denoted as $\text{NDPNet}\_{\textit{Cos}}$, for performing FSFGIC tasks on the Stanford Dogs and Stanford Cars datasets. Their corresponding results on the 5-way 1-shot and 5-way 5-shot tasks are presented in Table~\ref{t4}. It is observed that the classification performance of the $\text{NDPNet}\_{\textit{Cos}}$ is close to the classification performance of our proposed $\text{NDPNet}$. It can also be seen from Tables~\ref{t2} and~\ref{t4} that the classification performance of the $\text{NDPNet}\_{\textit{Cos}}$ is far better than that of the cosine similarity measure based DN4~\cite{li2020revisiting} method. This indicates that the non-linear data projection strategy effectively improve the ability to identify different samples.

Thirdly, we consider the classification performance of inner product measure based Gaussian kernel function learning network, which is named as $\text{GK\_FRaENet}$, after removing the designed non-linear data projection strategy. Based on the obtained descriptors ${\boldsymbol{q}}_{j}$ and ${\boldsymbol{s}}_{j}$ $(j=1,...,m)$ from the feature re-abstraction embedding network, the $\text{GK\_FRaENet}$ is designed as follows
\begin{equation}\begin{aligned}
\label{eq8}
K({\boldsymbol{q}}_{j},{\boldsymbol{s}}_{j})&=1-\text{exp}\bigg(-\frac{\Lambda({\boldsymbol{q}}_{j},{\boldsymbol{s}}_{j})}{2\delta^2}\bigg),\\
j&=1,...,m,
\end{aligned}\end{equation}
where $\delta$ is a scale factor. It is easy to verify that the function $K({\boldsymbol{q}}_{j},{\boldsymbol{s}}_{j})$ in Equation~(\ref{eq8}) satisfies the Mercer's theorem~\cite{mercer1909xvi}. It means that the function $K({\boldsymbol{q}}_{j},{\boldsymbol{s}}_{j})$ can be used for kernel learning which is utilized to compute the relationships between query images and support classes via Equations~(\ref{eq6}) and (\ref{eq7}). The $\text{GK\_FRaENet}$ is employed for performing FSFGIC tasks on the Stanford Dogs and Cars datasets. Their corresponding results on the 5-way 1-shot and 5-way 5-shot tasks are presented in Table~\ref{t4}. We can see from Table~\ref{t4} that our proposed $\text{NDPNet}$ achieves better classification performance than the $\text{GK\_FRaENet}$. This further validates that the proposed $\text{NDP-SM}$ can be more effectively making the distance between the samples within the same class smaller and the distance between samples from different classes larger.

%The reason is that compared with $\text{GK\_FRENet}$, our proposed $\text{NDPNet}$ can be more effectively to make the distance between the samples within the same class smaller and the distance between samples from different classes larger.

{\bf Influence of the $k$-nearest neighbors for FSFGIC.} The selection of parameter $k$ will affect the performance of classification. Here, we analyse the effect of the parameter $k$ in Equation~(\ref{eq5}) on the performance in performing FSFGIC tasks. 5-way 1-shot and 5-way 5-shot tasks are performed on the Stanford Dogs, Standford Cars, and CUB-200 datasets by using different $k$ ($k\in\{1,3,5\}$). Their corresponding results on the 5-way 1-shot and 5-way 5-shot tasks are presented in Table~\ref{t5}. It can be seen from Table~\ref{t5} that when $k$ equal to 1, our proposed method achieves the best classification performance. Thus, $k$-nearest neighbors with $k=1$ is recommended for our proposed architecture.

\section{Conclusions}
In this paper, we present a novel non-linear data projection network, $\text{NDPNet}$, which consists of a feature re-abstraction embedding module and a non-linear data projection based similarity measure module, for the challenging few-shot fine-grained image classification (FSFGIC). We emphasize and verify that the proposed NDPNet network is more effective for performing FSFGIC tasks. In this work, a feature re-abstraction embedding network is designed which is capable of representing a sample by a set of feature representations with detail re-enhanced semantic information. Furthermore, a novel non-linear data projection based similarity metric learning network is designed to make the distance between the samples within the same smaller and the distance between samples from different classes larger for performing FSFGIC. Our proposed method does not require extra supervision information and can be easily embedded into the episodic training mechanism for end-to-end training. Experiments on four fine-gained image classification benchmark datasets demonstrate that the presented $\text{NDPNet}$ method outperforms the state-of-the-art methods on various FSFGIC tasks.

%In this paper, we presented a taxonomy of the IFI extraction techniques for interest point detection. A comprehensive review on IFI extraction techniques for interest point detection is carried out. In addition, different issues of interest point detection methods are discussed. Furthermore, the main unresolved problems related to the existing IFI extraction techniques for interest point detection are identified and discussed. We also introduced the existing popular datasets and evaluation criteria. And then the performances for the eighteen most representative interest point detection methods are evaluated and discussed in terms of average recall rates under different camera positions and lighting changes, average repeatabilities under different image affine transformations, JPEG compressions, and noise degradations, and matching scores based on the VLBenchmarks with four different descriptors. Finally, the development trend for IFI extraction for interest point detection is elaborated.

%\ifCLASSOPTIONcompsoc
%  % The Computer Society usually uses the plural form
%  This work was supported by the Youth Innovation Team of Shaanxi Universities (China).
%\else
%  % regular IEEE prefers the singular form
%\fi

\ifCLASSOPTIONcaptionsoff
  \newpage
\fi

\renewcommand\refname{Reference}
\bibliographystyle{IEEEtran}
\bibliography{egbib}


%\vfill


\end{document}